\documentclass[journal]{IEEEtran}
\usepackage{amssymb}
\usepackage[subfigure]{graphfig}
\usepackage{amsmath}
\usepackage{amsfonts}
\usepackage{graphicx}
\usepackage{epsfig}
\usepackage{cases}
\usepackage{colortbl}
\usepackage{array,color}
\usepackage{algorithm}
\usepackage{algorithmic}
\usepackage{multirow}
\usepackage{color}

\usepackage{pifont}
\usepackage{multirow}

\usepackage{array}
\newcolumntype{I}{!{\vrule width 3pt}}
\newlength\savedwidth

\newlength\savewidth
\newcommand\shline{\noalign{\global\savewidth\arrayrulewidth
                            \global\arrayrulewidth 1.0pt}%
                   \hline
                   \noalign{\global\arrayrulewidth\savewidth}}

\ifCLASSINFOpdf
\else
\fi
\begin{document}
%
\title{Structure-Aware NeRF without Posed Camera via Epipolar Constraint}
%
%
%

\author{Shu~Chen,
        Yang~Zhang,
        Yaxin~Xu,
        and~Beiji~Zou
\thanks{S.~Chen, Y.~Zhang and Y.~Xu are with the School of Computer Science.School of Cyberspace Security, Xiangtan University, Xiangtan, China(e-mail: csu\_cs@163.com; 460086012@qq.com; 1253017164@qq.com.}
\thanks{B. Zou is with School of Computer Science and engineering, Central South University, Changsha, China (e-mail: bjzou@csu.edu.cn).}}
\maketitle

\begin{abstract}
The neural radiance field (NeRF) for realistic novel view synthesis requires camera poses to be pre-acquired by a structure-from-motion (SfM) approach. This two-stage strategy is not convenient to use and degrades the performance because the error in the pose extraction can propagate to the view synthesis. We integrate the pose extraction and view synthesis into a single end-to-end procedure so they can benefit from each other. For training NeRF models, only RGB images are given, without pre-known camera poses. The camera poses are obtained by the epipolar constraint in which the identical feature in different views has the same world coordinates transformed from the local camera coordinates according to the extracted poses. The epipolar constraint is jointly optimized with pixel color constraint. The poses are represented by a CNN-based deep network, whose input is the related frames. This joint optimization enables NeRF to be aware of the scene's structure that has an improved generalization performance. Extensive experiments on a variety of scenes demonstrate the effectiveness of the proposed approach. Code is available at {\color{blue}https://github.com/XTU-PR-LAB/SaNerf.}
\end{abstract}


%
\IEEEpeerreviewmaketitle

\section{Introduction}
%
%
%
%
\IEEEPARstart{V}{olumetric} neural rendering techniques have shown very promising results for novel-view synthesis of static scenes. Neural radiance fields (NeRF) \cite{Martinez2017} which implicitly represent a scene as a continuous five-dimensional (5D) function by a trained multi-layer perceptron (MLP) use volume rendering to synthesize a novel-view. Although NeRF and its variants have demonstrated unprecedented performance for view synthesis in a range of challenging scenes, most variants require the camera poses to be accurately estimated in advance. The camera poses are either directly accessible at training or extracted by a structure-from-motion (SfM) approach such as COLMAP \cite{Schonberger2016}. Accurate camera pose estimation is not a trivial task, and it can greatly affect the following NeRF.

To reduce the heavy dependence on having accurate camera pose information, Wang et al. \cite{Wang2021} jointly optimized both the parameters of the NeRF model and the camera parameters by minimizing the photometric reconstruction error. However, the employed loss focuses on view synthesis, meaning the model could fall into local minima where the optimized camera parameters are sub-optimal, resulting in unrealistic synthesized images. Recently, GNeRF \cite{Meng2021} attempted to train NeRF for complex scenarios without known camera poses by using generative adversarial networks (GAN); however, these networks rely on a camera sampling distribution not far from the true distribution. In contrast, our approach does not assume a certain prior and can be optimized from randomly initialized camera poses.

NeRF is able to create photo-realistic novel views of scenes with complex geometry and appearance because it relies on enough images to overcome the inherent ambiguity of the correspondence \cite{Roessle2022}, just as the direct methods in simultaneous localization and mapping (SLAM). For example, LSD-SLAM \cite{Engel2014} reconstructs the structure and appearance of scenes by photometric consistent constraint. In contrast to LSD-SLAM, NeRF is an indirect method which implicitly constrains the structure of scenes by synthesizing different views. Its performance drops significantly when the input views are sparse because it often finds a degenerate solution to its image reconstruction objective. Some researchers \cite{Roessle2022, Wei2021} have addressed the issue by employing the estimated depth as prior to monitor the sampling process of volume rendering. However, this kind of solution still treats structure reconstruction and view synthesis as two separated processes that cannot benefit from each other. In the traditional SfM or SLAM, the structure and appearance of the scenes are simultaneously reconstructed by optimizing an objective function. Inspired by the joint optimization approach that comes from bundle adjustment in classical SfMs, we propose a structure-aware NeRF without posed camera approach (SaNeRF).

To benefit from the strengths of structure reconstruction and view synthesis, in this work, we jointly optimize the camera's extrinsic parameters and scene representation during the training. Our approach is to incorporate a camera extrinsic parameters optimization solver in the MLP architecture, which implicitly represents the appearance of scenes, and allow the solver to participate in guiding the updates of the MLP parameters, thereby realizing end-to-end learnable model training. Specifically, we first extracted the SIFT \cite{Lowe2004} pairs in two views. Then, the global 3D coordinates of each SIFT and its match in these views were optimized according to the epipolor constraint that each SIFT and its correspondence must have the same 3D global coordinates. The epipolor constraint enables NeRF to be aware of the scene's structure. The global 3D coordinates of each SIFT were obtained by transforming the camera coordinates into the global coordinate system according to the camera's extrinsic parameters. The camera extrinsic parameters are extracted by a CNN in which the input is the related images.
To sum up, our main contributions include:

1) The requirement that the camera poses must be pre-known in NeRF is addressed by the joint optimization of the camera's extrinsic parameters and scene representation; our approach only utilizes randomly initialized poses for complex scenarios.

2) The employed epipolor constraint enables NeRF to be aware of the scene's structure so that the generalization ability of NeRF is improved.

3) The camera's pose extraction and scene representation are integrated in an end-to-end manner, which frees NeRF from the inaccurate estimation of camera poses, and can benefit from each other.

\section{Related Work}
\subsection{Scene Representations}
Scenes are traditionally represented as: grids of voxels \cite{Wu2015}, meshes \cite{Litany2018} and point clouds \cite{Qi2017}. Conventional signal representations are usually discrete, so they are indifferentiable.  In contrast, implicit neural representations represent a scene as a continuous function that maps the 3D coordinates to whatever is at that coordinate (for texture, an RGB color; for shape, a density value) \cite{Park2019, Mescheder2019}. Park et al. \cite{Park2019} introduced a learned continuous Signed Distance Function (SDF) representation of a class of shapes that enables high quality shape representation, interpolation and completion from partial and noisy 3D input data. Some works \cite{Atzmon2020, Gropp2020, Chibane2020} have improved it by learning SDFs from raw data (\emph{i.e.}, without ground-truth signed distance values). Besides implicit neural representations of geometry, geometry and appearance are simultaneously represented by a neural network \cite{Saito2019, Oechsle2019}. However, these methods need costly 3D supervision.

To achieve self-supervised implicit neural representation learning, 3D scenes are represented as 3D-structured neural scene representations which are formulated from 2D images \cite{Martinez2017, Liu2019, Niemeyer2020, Lin2020, Yariv2020, Kaviani2018}. Sitzmann et al. \cite{Sitzmann2019} proposed Scene Representation Networks (SRNs), which represent scenes as continuous functions that map world coordinates to a feature representation of local scene properties. Niemeyer et al. \cite{Niemeyer2020} replaced long short-term memory (LSTM)-based ray-marcher in SRNs with a fully-connected neural network, enabling easy extraction of the final 3D geometry. NeRF [1]\cite{Martinez2017} combines an implicit neural model with volume rendering for novel view synthesis of complex scenes. Mip-NeRF \cite{Barron2021} extends NeRF to represent the scene at a continuously valued scale by efficiently rendering anti-aliased conical frustums instead of rays. NeRF has inspired many subsequent works: fast inference \cite{Lindell2021}, deformable \cite{Pumarola2021}, scalable \cite{Tancik2022}, and generalization \cite{Schwarz2020, Trevithick2021, Xu2022}. However, they all assume accurate camera poses as a prerequisite.

\subsection{Structure-aware View Synthesis}
A few works have proposed incorporating depth observations into NeRF reconstruction. Nerfing-MVS \cite{Wei2021} used the learning-based priors to guide the optimization process of NeRF; the priors were obtained by training a monocular depth network based on the sparse SfM estimation. Roessle et al. \cite{Roessle2022} employed a depth completion network to convert these sparse points obtained by the SfM into dense depth maps, then leveraged dense depth priors to constrain the NeRF optimization. DS-NeRF \cite{Deng2022} proposed a similar approach, but the approach considered the reliability of estimated depth in the depth constraint. Because the depth prior estimation and NeRF are two separated processes in which small camera calibration errors may impede photorealistic synthesis \cite{Zhang2020}, they cannot benefit from each other. In comparison, our approach integrates the sparse 3D structure estimation and view synthesis into an end-to-end manner.

Several works exploit generative 3D models for 3D-aware image synthesis \cite{Liao2020, Niemeyer2021}. Some works have employed explicit methods that either generate a voxelized 3D model \cite{Henzler2019} or learn an abstract 3D feature representation \cite{Nguyen-Phuoc2019}. Due to the learned neural projection function, it is prone to create discretization artifacts or degrades view-consistency of the generated images. In comparison, implicit methods \cite{Meng2021, Schwarz2020, Niemeyer2021, Nguyen-Phuoc2020} apply rotation to a 3D latent feature vector to generate transformed images through a MLP. Schwarz et al. \cite{Schwarz2020} proposed a generative model for radiance fields for novel view synthesis of a single scene by introducing a multiscale patch-based discriminator. Niemeyer et al. proposed \cite{Niemeyer2021} a controllable image synthesis pipeline which can be trained from raw image collections without additional supervision. However, all these approaches are only able to learn models for single objects or static scenes.

\subsection{Camera Pose Estimation}
Classical methods for object pose estimation address the task by template matching with a specific 3D model \cite{Chen2016, Ferrari2006}. However, the employed model is commonly inaccurate in that the performance of object pose estimation is significantly affected. In contrast, SfM \cite{Schonberger2016, Chen2022, Wilson2014} is able to simultaneously recover the 3D structure of scenes and camera poses. In SfM, the initial camera poses are estimated in two consecutive steps: (1) features such as SIFT are extracted from each image and the identical features belonging to the consecutive image frames are matched; (2) the camera poses are inferred by five-point \cite{Nister2004} or eight-point \cite{Hartley1995} algorithms, according to the matches. The extracted camera poses are further refined by a bundle adjustment (BA) optimization. Some deep learning-based methods \cite{Kendall2015, Sundermeyer2018, Fridovich-Keil2021} attempt to regress the camera pose directly from 2D images without the need of tracking. However, non-linearity of the rotation space limits the generalization ability of DNN-based method. In contrast, keypoint-based approaches \cite{Peng2019, Liu2020} detect the keypoints of objects and solve for pose using the PnP-RANSAC algorithm. By inverting a trained model \cite{Meng2021, Chen2021}, NeRF has also been explored for pose estimation.

\section{Preliminary}
NeRF is a continuous 5D function mapping a 3D location ${\bf{x}} = \left( {x,y,z} \right)$ and 2D viewing direction $\left( {\theta ,\varphi } \right)$ to an emitted color ${\bf{c}} = \left( {r,g,b} \right)$ and volume density $\sigma $. The continuous function is implicitly represented by a MLP where the weights of the MLP are optimized to reconstruct a set of input images of a particular scene.
Given \emph{n} training images and the corresponding camera poses, the NeRF is optimized according to a photometric loss as
\begin{equation}\label{eq1}
L = \frac{1}{n}\sum\limits_{i = 1}^n {\left\| {{I_i} - {{\hat I}_i}} \right\|_2^2} ,
\end{equation}
where ${I_i}$ is the ground-truth color of image \emph{i} and ${\hat I_i}$ is the corresponding synthesized image by volume rendering.

For each pixel of ${\hat I_i}$, casting a ray ${\bf{r}}\left( t \right) = {\bf{o}} + t{\bf{d}},\;{\rm{ }}{\bf{o}} \in {\bf{R}^3},\;{\rm{ }}{\bf{d}} \in {S^2},\;{\rm{ }}t \in \left[ {{t_n},{t_f}} \right]$ from the camera center ${\bf{o}}$ through the pixel along direction ${\bf{d}}$, and its color renders as
\begin{equation}\label{eq2}
{{\bf{\hat c}}_\theta }\left( {\bf{r}} \right) = \int_{{t_n}}^{{t_f}} {T\left( t \right){\sigma _\theta }\left( {{\bf{r}}\left( t \right)} \right)} {{\bf{c}}_\theta }\left( {{\bf{r}}\left( t \right),{\bf{d}}} \right)dt,
\end{equation}
where $T\left( t \right) = \exp \left( { - \int_{{t_n}}^t {{\sigma _\theta }\left( {{\bf{r}}\left( s \right)} \right)ds} } \right),$ and ${\sigma _\theta }\left(  \cdot  \right)$ and ${{\bf{c}}_\theta }\left( { \cdot , \cdot } \right)$ indicate the volume density and color prediction of the radiance field, respectively.

There are two drawbacks in NeRF:

(1)	Precomputed camera parameters are required to train a NeRF model. As pointed out by \cite{Zhang2020} that small camera calibration errors may impede photorealistic synthesis.

(2)	NeRF's performance drops significantly if the number of input views is sparse \cite{Niemeyer2022}. NeRF learns to reconstruct the input views perfectly. Novel views may degenerate because the model is not biased towards learning a 3D-consistent solution in such a sparse input scenario.

We explain the second drawback as follows:

Given an opaque voxel (in the surface) in the scene, the color of this voxel projected onto one view is $p(v)$; ${\sigma _\theta }\left(  \cdot  \right)$ must satisfy the following constraint, as shown in Fig. 1(a).
\begin{equation}\label{eq3}
\begin{aligned}
\begin{split}
&{\bf{c}}\left( {{\bf{r}}{\rm{(}}t{\rm{)}}} \right) = p(v)\;{\rm{ and }}\;\sigma \left( {{\bf{r}}{\rm{(}}t{\rm{)}}} \right) = 1,{\rm{ }}\\
&{\rm{while }}\;{\bf{r}}{\rm{(}}t{\rm{) \;is \;the \;location \;of \;the \;voxel,}}\\
&\sigma \left( {{\bf{r}}{\rm{(}}t{\rm{)}}} \right) = 0,\;{\rm{ otherwise}}{\rm{.}}
\end{split}
\end{aligned}
\end{equation}

However, according to (2), there are many solutions with different $\sigma \left( {{\bf{r}}{\rm{(}}t{\rm{)}}} \right)$ and ${\bf{c}}\left( {{\bf{r}}{\rm{(}}t{\rm{)}}} \right)$ that make ${{\bf{\hat c}}_\theta }\left( {\bf{r}} \right)$ render as $p(v)$. Fig. 1(b) demonstrates one choice in which the pixel value (the green dot) is represented by the mixture of projections of two voxels (yellow dot and blue dot), and this is called shape-radiance ambiguity \cite{Zhang2020}. Fig. 1(c) shows that introducing more constraints can eliminate this ambiguity to some extent. In this situation, if the opaque voxel is projected to another view, that the photometric loss in the other view makes Fig. 1(b) impossible. Therefore, NeRF requires dense inputs to introduce more constraints to avoid degenerate solutions.

Our approach jointly optimizes the camera's parameters and scene representations during the training to address the aforementioned drawbacks. The combination strategy enables them in a mutually reinforcing manner.

\begin{figure*}
  \centering
  \includegraphics[width = 5.0in]{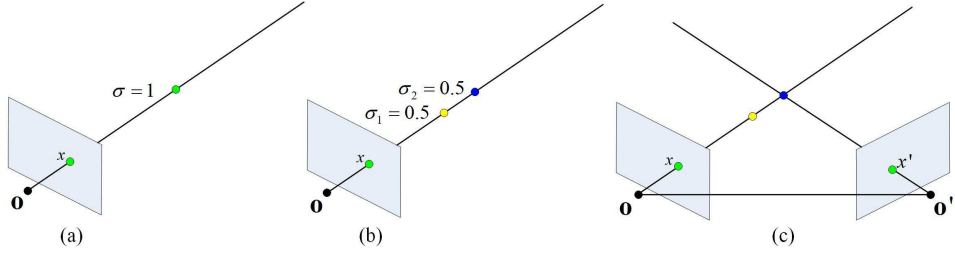}\\
  \caption{(a) The correct solution; (b) one incorrect solution; (c) projecting the voxel to another view makes (b) impossible. $x$ and $x'$ are the two camera centers, respectively. ${\bf{o}}$ and ${\bf{o}}'$ are the projections of the voxel projecting onto two views, respectively. $\sigma $ is the volume density.
}\label{f1}
\end{figure*}

\section{SaNeRF}
\subsection{Overview}
Fig. 2 shows the pipeline of our SaNeRF framework for novel views synthesis without a posed camera. Given a dataset with \emph{n} images, we first extract the sift matches between any three images selected from the dataset. Then an image from n images is determined as the reference image that its camera coordinate system is the world coordinate system, according to how many images it matched and the number of matches. We input any triple images in which one image is the reference image into a pose network to estimate the extrinsic parameters of the other two images related to the reference image. The estimated extrinsic parameters of the other images and the extrinsic parameters of the reference image ($\left\{ {{\bf{I}},0} \right\}$,  ${\bf{I}}$ is the identity matrix and $0$ is the zero vector), are then input into the NeRF to synthesize their images. In NeRF, the input poses are used to transform the rays in the camera coordinate system into the world coordinate system by
\begin{equation}\label{eq4}
{\bf{d}}' = {\bf{Rd}};\;{\rm{ }}{\bf{o}}' = {\bf{t}},
\end{equation}
where $\left( {{\bf{R}},{\bf{t}}} \right)$ are the estimated extrinsic parameters; ${\bf{d}}'$ and ${\bf{o}}'$ are the viewing direction and start point of a ray after transformation, respectively; ${\bf{d}}$ is the viewing direction of the ray before transformation, and the start point of the ray before transformation is $\left( {0,0,0} \right)$.

In addition, the 3D coordinates of the sift features in each image are estimated according to the volume densities output from the neural radiance fields, and formulated as a weighted sum of all samples volume densities ${\sigma _i}$ along the ray, defined as
\begin{equation}\label{eq5}
{{\bf{x}}_s} = \sum\limits_{i = 1}^{{N_c}} {{w_i}\left( {{\bf{o}}' + {t_i}{\bf{d}}'} \right)} ,\;{\rm{ }}{w_i} = {T_i}\left( {1 - \exp \left( { - {\sigma _i}{\delta _i}} \right)} \right).
\end{equation}

Besides the photometric reconstruction loss (${L_{pm}}$) introduced in NeRF, positions of matched features loss (${L_{3D}}$) is employed in SaNeRF to optimize the pose network, according to the multi-view geometry in which the identical features in different views have the same 3D world coordinates.
\begin{figure*}
  \centering
  \includegraphics[width = 6.2in]{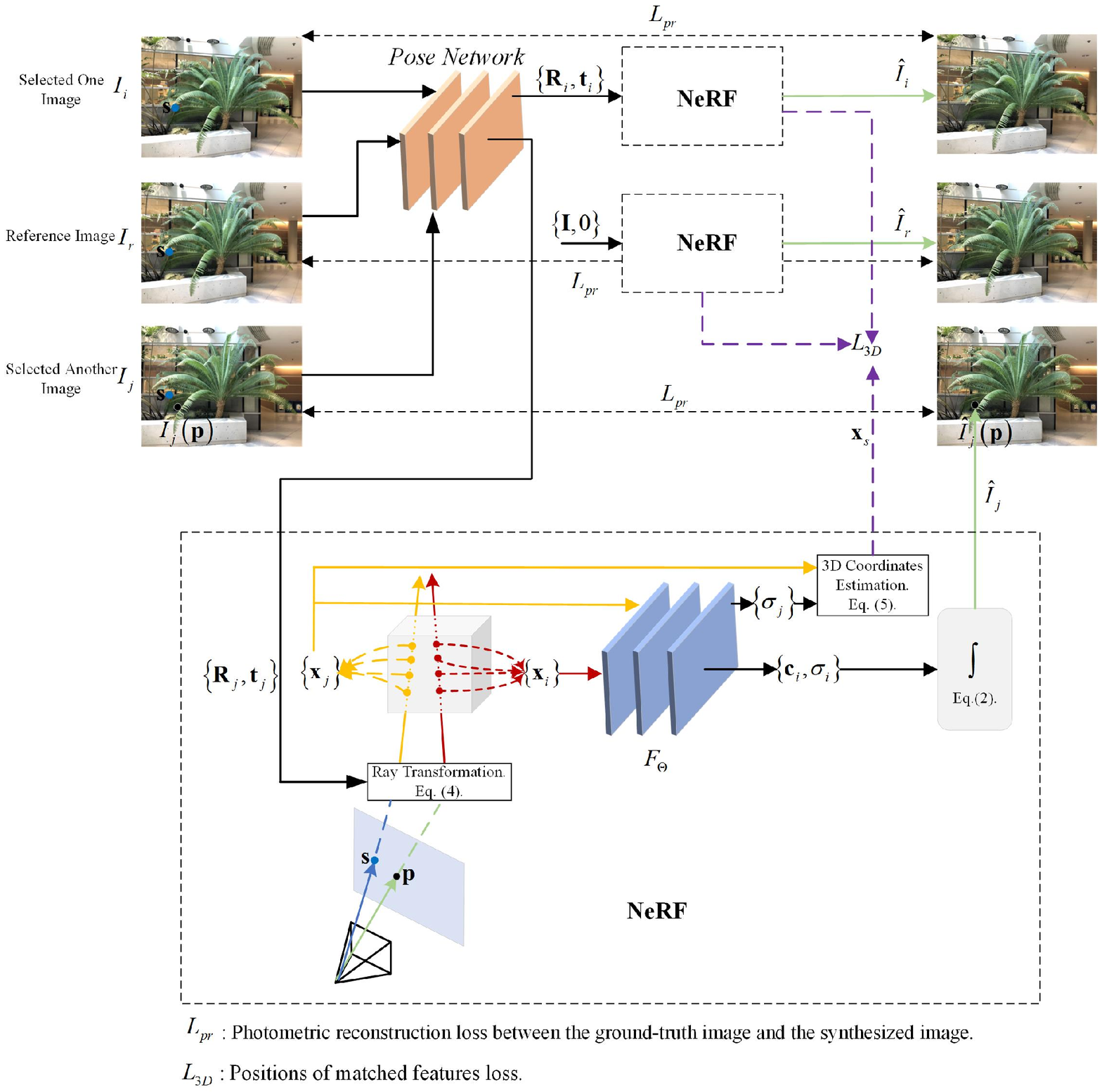}\\
  \caption{Overview of SaNeRF. The blue points in the selected images are the matched features of the blue point in the reference image. ${F_\Theta }$ is the neural radiance field which represented by a MLP. ${\bf{s}}$ represents the location of one matched SIFT correspondence in the image.
}\label{f2}
\end{figure*}

\subsection{Pre-Processing}
For each image in the dataset, we established a set of three images in which an image is assumed to be the reference image and the other two images were selected from the dataset without replacement. We extracted SIFT features in each image and obtained the matched SIFT correspondences between them. If the number of matched SIFT correspondences was above three then the other two images were denoted as the matched images of that image. By doing this, we were able to determine the number of matched images of each image and the number of matched correspondences.

The reference image was the image with the maximum number of matched images and the maximum number of matched correspondences.

\subsection{Pose Network}
Following \cite{Zhou2017, Chen2022-2}, we employ fully convolutional architecture to model the pose network. The input to the pose network is the reference image concatenated with the other two images (along the color channels), and the outputs are the relative poses between the reference image and each of the other two images. The network consists of seven stride-2 convolutions. Except for the kernel sizes of the first and the second which are seven and five, respectively, all other convolutions are three. The last convolutional layer is followed by a $1 \times 1$ convolution with $6 \times \left( {N - 1} \right)$ output channels (corresponding to three Euler angles and 3D translation for each image). Finally, global average pooling is applied to aggregate predictions at all spatial locations.

\subsection{Training loss}
We extend the loss defined in (1) to include the photometric reconstruction error of SIFT features, and the final photometric reconstruction loss is defined as
\begin{equation}\label{eq6}
{L_{pr}} = L + \frac{1}{{nm}}\sum\limits_{i = 1}^n {\sum\limits_{j = 1}^m {\left\| {{\bf{c}}_i^k - {\bf{\hat c}}_i^k} \right\|_2^2} } ,
\end{equation}
where ${\bf{c}}_i^k$ is the ground-truth color of the matched SIFT feature \emph{k} in the image \emph{i} which is obtained by the bilinear interpolation at the location of the SIFT \emph{k} from the image \emph{i}. ${\bf{\hat c}}_i^k$ is the corresponding rendered color from NeRF.

At the pre-processing, we extracted SIFT features at each frame and matched them between three images. We denote the 3D coordinates of matched correspondences as $\left\{ {\left[ {{\bf{x}}_r^k,{\bf{x}}_i^k,{\bf{x}}_j^k} \right]\left| {k = 1, \cdots ,m} \right.} \right\}$, and the positions of matched features loss is defined as
\begin{small}
\begin{equation}\label{eq7}
{L_{3D}} = \frac{1}{{nml}}\sum\limits_{i = 1}^n {\sum\limits_{j = 1}^m {\sum\limits_{k = 1}^l {\left\| {{\bf{x}}_r^k - {\bf{x}}_i^k} \right\|_2^2 + \left\| {{\bf{x}}_r^k - {\bf{x}}_j^k} \right\|_2^2 + \left\| {{\bf{x}}_i^k - {\bf{x}}_j^k} \right\|_2^2} } } ,
\end{equation}
\end{small}
where ${\bf{x}}_r^k$, ${\bf{x}}_i^k$, and ${\bf{x}}_j^k$ are the estimated 3D coordinates of the matched sift features \emph{k} in the reference image and another two images according to (5), respectively.

The total loss is defined as a combination of two items; each item is controlled by a factor.
\begin{equation}\label{eq8}
{L_{total}} = \alpha {L_{pr}} + \beta {L_{3D}}.
\end{equation}

\section{Experiment}
\subsection{Experimental Details}
We used the publicly available Pytorch framework to implement our code and evaluate the performance of our system on the LLFF-NeRF \cite{Martinez2017, Mildenhall2019} and ScanNet \cite{Dai2017} datasets.

\textbf{Datasets:} The LLFF Dataset consists of eight scenes captured with a handheld cellphone, with 20-62 images each. The resolution of each image in the dataset is $4032 \times 3024$. Due to the limited capacity of NVIDIA RTX 2080Ti, we downsized each image 1/8 scale to $504 \times 378$ dimensions in pixels, and held out 1/8 of these as the test set for novel view synthesis.

Following the experimental setup in \cite{Mildenhall2019}, we selected eight scenes in the ScanNet dataset to evaluate our method. In each scene, 40 images were selected to cover a local region, and all images were resized to $648 \times 484$. Similar to NeRF \cite{Martinez2017}, we held out 1/8 of these as the test set for novel view synthesis.

\textbf{Training details:} Differing from the network architecture of the original NeRF \cite{Martinez2017}, we employed a similar MLP in NeRF \cite{Martinez2017} for each of the emitted RGB radiance and volume density predictions. The hierarchical sampling strategy in NeRF \cite{Martinez2017} was adopted and numbers of sampled points of both coarse sampling and importance sampling were set to 64. We optimized our model with the Adam \cite{Kingma2014} optimizer with ${\beta _1} = 0.9,{\beta _2} = 0.999$ and the learning rate was 5e-4. The loss weights were set as $\alpha  = 1.0$ and $\beta  = 1.0$. One NVIDIA RTX 2080Ti was used for training and testing and SaNeRF took about 20-60 hours to train a single scene of the LLFF dataset.

\textbf{Evaluation Metrics:} We employed two kinds of metrics to evaluate the proposed approach: For the quality of novel view rendering measurement, we used the common metrics: Peak Signal-to-Noise Ratio (PSNR), Structural Similarity Index Measure (SSIM) \cite{Wang2004} and Learned Perceptual Image Patch Similarity (LPIPS) \cite{Zhang2018}; For the accuracy of the optimized camera parameters evaluation, we computed the Absolute Trajectory Error (ATE) \cite{Sturm2012}, which first aligns two sets of pose trajectories globally using a similarity transformation Sim(3) and reports the absolute distance between two translation vectors. Since the ground-truth was not available, we ran COLMAP \cite{Schonberger2016} to obtain the extrinsic parameters of each camera, and used ATE to evaluate the accuracy by computing the difference between our optimized camera and the estimations from COLMAP.

\subsection{Comparison with State-of-the-Art}
\subsubsection{Novel View Synthesis Comparison}
For evaluation, we need to estimate the camera poses of the test view images. We trained our approach on each scene two times. The first time, we trained SaNeRF on all images in the scene, in order to estimate the cameras' poses in the test images; The second time, we trained SaNeRF only on the training images to evaluate the quality of the novel view rendering from the cameras in the test images in which the poses of the test images were obtained from the first training.

\textbf{Results on LLFF-NeRF dataset.}

We compared SaNeRF to NeRF \cite{Martinez2017} as well as recent and concurrent work include: Plenoxels \cite{Fridovich-Keil2021}, BARF \cite{Lin2021}, NeRF-\,- \cite{Wang2021} and TensoRF \cite{Gropp2020}. The quantitative comparisons for novel view synthesis are shown in Table 1, and the visual results are illustrated in Fig. 3. Our method achieved the best performance measured by PSNR and SSIM in scenes (\emph{fern}, \emph{leaves} and \emph{orchids}) because the rich texture in these scenes can guarantee enough reliable keypoints for precise camera poses estimation. As compared to BARF \cite{Lin2021}, which trains NeRF from imperfect camera poses, our approach outperformed it in five scenes. As compared to NeRF-\,- without known camera parameters, our approach outperformed it in almost all of the scenes except for \emph{scene T-Rex}. Our approach achieved the best performance in six scenes on LPIPS because LPIPS is free from the errors that are caused by the inaccurately estimated camera poses.

\begin{figure}
  \centering
  \includegraphics[width = 3.4in]{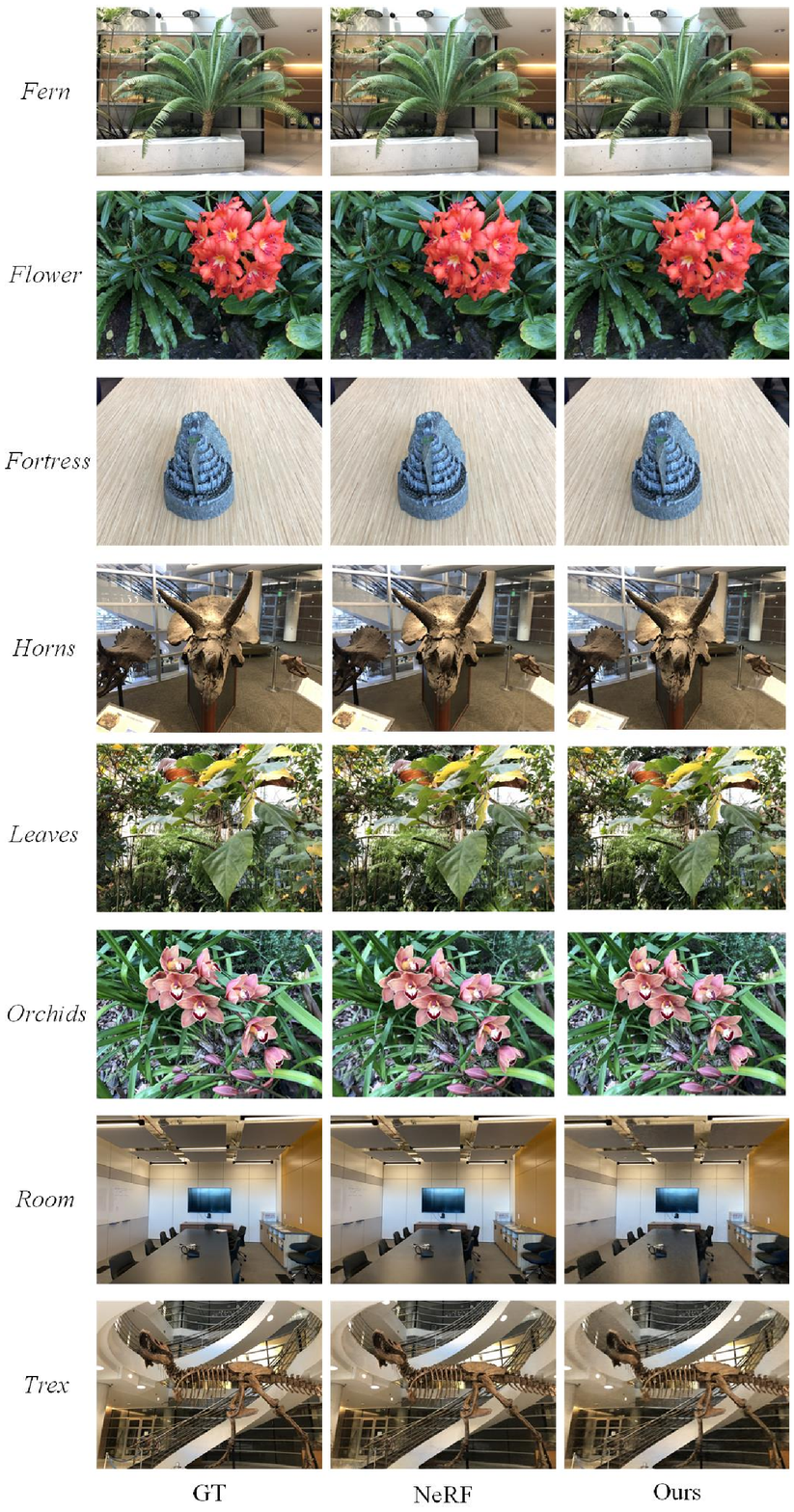}\\
  \caption{Qualitative comparison between our SaNeRF with unknown cameras and other approaches on the LLFF-NeRF dataset.
}\label{f3}
\end{figure}

\begin{table*}[!htb]
\renewcommand\arraystretch{1.2}
\footnotesize
\begin{center}
\begin{tabular}{p{3.2cm}|p{0.8cm}p{0.8cm}p{0.8cm}p{0.8cm}p{0.8cm}p{0.8cm}p{0.8cm}p{0.8cm}}
\shline
Methods & \emph{Room} & \emph{Fern} & \emph{Leaves} & \emph{Fortress} & \emph{Orchids} & \emph{Flower} & \emph{T-Rex} & \emph{Horns} \\
\hline
\multicolumn{9}{l}{\textbf{PSNR}$\uparrow$} \\
\hline
NeRF \cite{Martinez2017} ECCV20 & 32.70 &	25.17 &	20.92 &	31.16 &	20.36 &	27.40 &	26.80 &	27.45 \\
Plenoxels \cite{Fridovich-Keil2021} arXiv21 &	30.22 &	25.46 &	21.41 &	31.09 &	20.24 &	27.83 &	26.48 &	27.58 \\
TensoRF \cite{Gropp2020} ECCV22 &	\textbf{32.35} &	25.27 &	21.30 &	\textbf{31.36} &	19.87 &	\textbf{28.60} &	\textbf{26.97} &	\textbf{28.14} \\
\hline
BARF \cite{Lin2021} ICCV21 &	31.95 &	23.79 &	18.78 &	29.08 &	19.45 &	23.37 &	22.55 &	22.78 \\
NeRF-\,- \cite{Wang2021} arXiv21 &	25.73 &	21.83 &	18.73 &	26.55 &	16.50 &	25.34 &	22.49 &	24.35 \\
Ours &	28.09 &	\textbf{25.95} &	\textbf{21.87} &	27.51 &	\textbf{21.40} &	26.98 &	20.15 &	26.29 \\
\hline
\multicolumn{9}{l}{\textbf{SSIM}$\uparrow$} \\
\hline
NeRF \cite{Martinez2017} ECCV20 &	0.948 &	0.792 &	0.690 &	0.881 &	0.641 &	0.827 &	0.880 &	0.828 \\
Plenoxels \cite{Fridovich-Keil2021} arXiv21 &	0.937 &	0.832 &	0.760 &	0.885 &	0.687 &	0.862 &	0.890 &	0.857 \\
TensoRF \cite{Gropp2020} ECCV22	& \textbf{0.952} &	0.814 &	0.752 &	\textbf{0.897} &	0.649 &	\textbf{0.871} &	\textbf{0.900} &	\textbf{0.877} \\
\hline
BARF \cite{Lin2021} ICCV21 &	0.940 &	0.710 &	0.537 &	0.823 &	0.574 &	0.698 &	0.767 &	0.727 \\
NeRF-\,- \cite{Wang2021} arXiv21 &	0.83 &	0.62 &	0.52 &	0.67 &	0.38 &	0.71 &	0.72 &	0.63 \\
Ours &	0.817 &	\textbf{0.837} &	\textbf{0.783} &	0.806 &	\textbf{0.747} &	0.858 &	0.702 &	0.857 \\
\hline
\multicolumn{9}{l}{\textbf{LPIPS $_vgg$}$\downarrow$} \\
\hline
NeRF \cite{Martinez2017} ECCV20 &	0.178 &	0.280 &	0.316 &	0.171 &	0.321 &	0.219 &	0.249 &	0.268 \\
Plenoxels \cite{Fridovich-Keil2021} arXiv21 &	0.192 &	0.224 &	0.198 &	0.180 &	0.242 &	0.179 &	0.238 &	0.231 \\
TensoRF \cite{Gropp2020} ECCV22 &	0.167 &	0.237 &	0.217 &	0.148 &	0.278 &	0.169 &	0.221 &	0.196 \\
\hline
BARF \cite{Lin2021} ICCV21 &	\textbf{0.099} &	0.311 &	0.353 &	0.132 &	0.291 &	0.211 &	\textbf{0.206} &	0.298 \\
NeRF-\,- \cite{Wang2021} arXiv21 &	0.44 &	0.49 &	0.47 &	0.44 &	0.56 &	0.37 &	0.44 &	0.49 \\
Ours &	0.281 &	\textbf{0.098} &	\textbf{0.117} &	\textbf{0.064} &	\textbf{0.102} &	\textbf{0.065} &	0.358 &	\textbf{0.091} \\
\shline

\end{tabular}
\end{center}
\caption{Quantitative comparisons for novel view synthesis on LLFF-NeRF dataset.}
\end{table*}

\textbf{Results on ScanNet dataset.}

Table 2 shows the quantitative comparisons for novel view synthesis on the ScanNet dataset. The qualitative results are illustrated in Fig. 4. Our approach achieved the best performance in scenes (\emph{scene 0079}, \emph{scene 0158} and \emph{scene 0553}). With other challenging scenes that did not have enough keypoints for pose estimation, our approach failed to synthesis good results. Our approach outperformed the original NeRF \cite{Martinez2017} in five scenes which benefitted from the joint optimization.

\begin{figure}
  \centering
  \includegraphics[width = 3.4in]{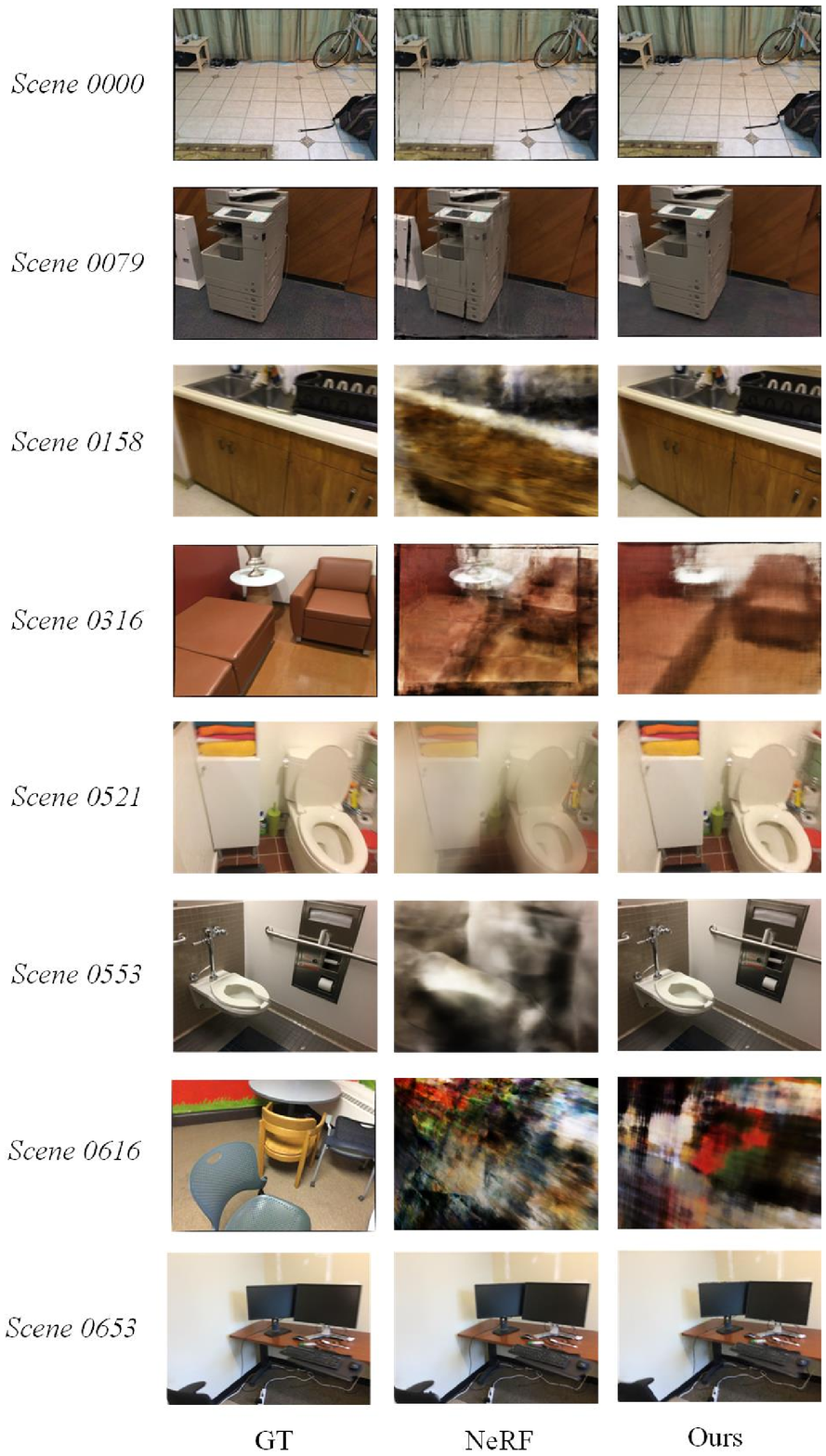}\\
  \caption{Qualitative comparison between our SaNeRF with unknown cameras and other approaches on the ScanNet dataset.
}\label{f4}
\end{figure}

\begin{table*}[!htb]
\renewcommand\arraystretch{1.2}
\footnotesize
\begin{center}
\begin{tabular}{p{3.2cm}|p{0.8cm}p{0.8cm}p{0.8cm}p{0.8cm}p{0.8cm}p{0.8cm}p{0.8cm}p{0.8cm}}
\shline
Methods & \emph{Scene 0000} & \emph{Scene 0079} & \emph{Scene 0158} & \emph{Scene 0316} & \emph{Scene 0521} & \emph{Scene 0553} & \emph{Scene 0616} & \emph{Scene 0653} \\
\hline
\multicolumn{9}{l}{\textbf{PSNR}$\uparrow$} \\
\hline
NSVF \cite{Liu2020-2} NeurIPS20	& \textbf{23.36}	& 26.88	& 31.98	& \textbf{22.29}	& 27.73	& 31.15	& 15.71	& 28.95 \\
SVS \cite{Riegler2021} CVPR21	& 21.39	& 25.18	& 29.43	& 20.63	& 27.97	& 30.95	& \textbf{21.38}	& 27.91 \\
\hline
NeRF \cite{Martinez2017} ECCV20	& 18.75	& 25.48	& 29.19	& 17.09	& 24.41	& 30.76	& 15.76	& 30.89 \\
NerfingMVS \cite{Wei2021} ICCV21	& 22.10	& 27.27	& 30.55	& 20.88	& \textbf{28.07}	& 32.56	& 18.07	& \textbf{31.43} \\
Ours	& 20.22	& \textbf{28.65}	& \textbf{33.12}	& 16.06	& 25.89	& \textbf{33.69}	& 14.56	& 28.55 \\
\hline
\multicolumn{9}{l}{\textbf{SSIM}$\uparrow$} \\
\hline
NSVF \cite{Liu2020-2} NeurIPS20	& 0.823	& 0.887	& 0.951	& 0.917	& 0.892	& 0.947	& 0.704	& 0.929 \\
SVS \cite{Riegler2021} CVPR21	& \textbf{0.914}	& 0.923	& 0.953	& \textbf{0.941}	& \textbf{0.924}	& 0.968	& \textbf{0.899}	& 0.965 \\
\hline
NeRF \cite{Martinez2017} ECCV20	& 0.751	& 0.896	& 0.928	& 0.828	& 0.871	& 0.950	& 0.699	& 0.953 \\
NerfingMVS \cite{Wei2021} ICCV21	& 0.880	& 0.916	& 0.948	& 0.899	& 0.901	& 0.965	& 0.748	& \textbf{0.964} \\
Ours	& 0.638	& \textbf{0.917}	& \textbf{0.955}	& 0.716	& 0.783	& \textbf{0.969}	& 0.598	& 0.899 \\
\shline

\end{tabular}
\end{center}
\caption{Quantitative comparisons for novel view synthesis on ScanNet dataset.}
\end{table*}

\subsubsection{Extrinsic Parameters Comparison}

We evaluated the accuracy of the camera parameter estimation on the LLFF-NeRF dataset. Since the ground-truth camera parameters for these sequences are not accessible, we ran COLMAP \cite{Schonberger2016} on all images to obtain camera parameters as references, and report the difference between our predicted camera parameters and theirs on the training images.

Table 3 shows the ATE comparison on the translation between our approach and other approaches. We compared our optimized camera trajectories with the ones estimated from COLMAP, aligned using ATE in Fig. 5. From Table 3 and Fig. 5, we notice that the camera poses obtained from our approach are close to those estimated from COLMAP, confirming the effectiveness of the joint optimization strategy.

To better understand the optimization process, we provide a visualization of the camera poses at various training epochs for the scene flower from the LLFF-NeRF dataset in Fig. 6. The pose estimations are randomly initialized at the beginning, and converged after about 100,000 epochs, subject to a similarity transformation between predicted camera parameters and those estimated from COLMAP.

\begin{figure*}
  \centering
  \includegraphics[width = 6.5in]{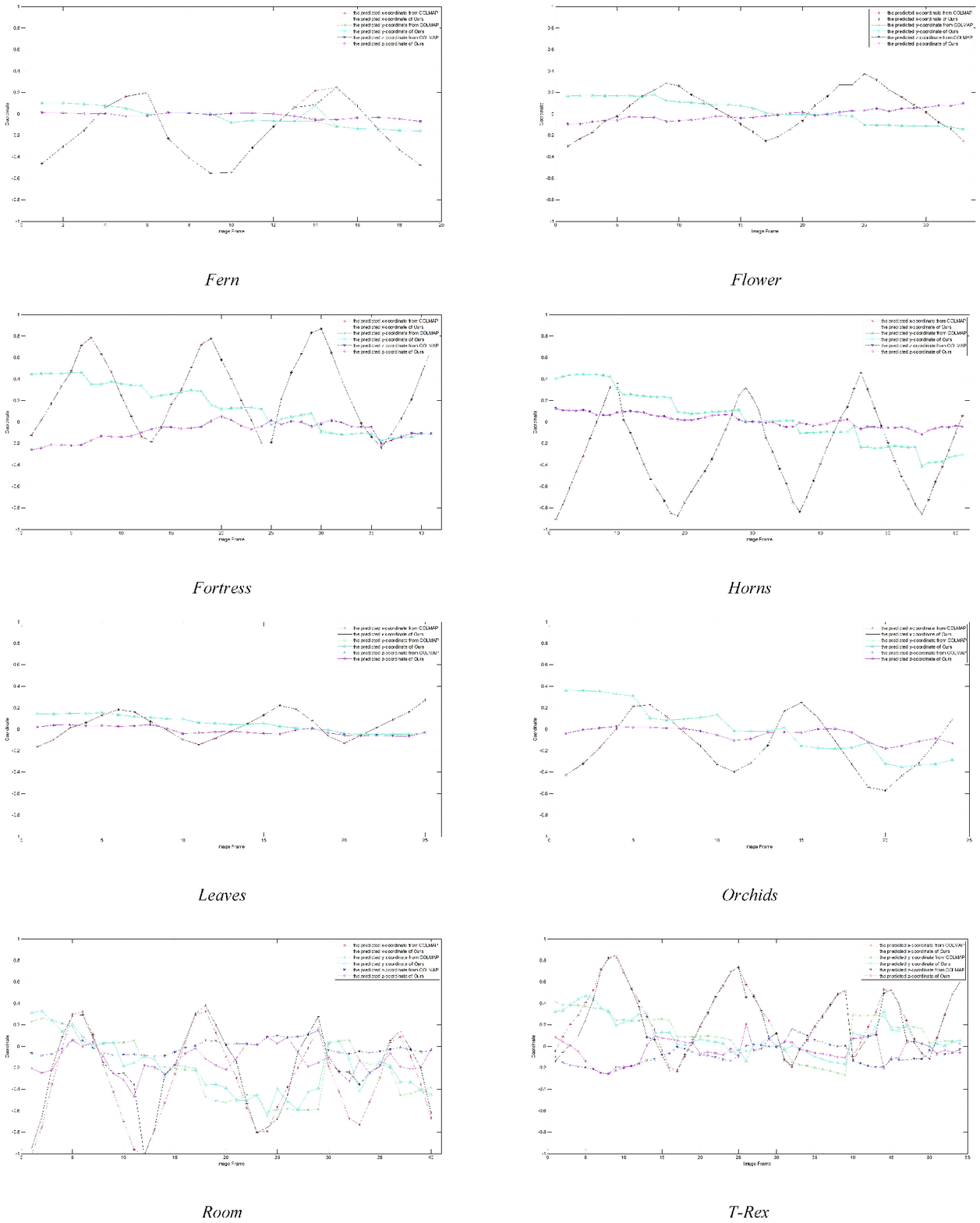}\\
  \caption{Comparison of camera trajectories between the optimized translations and the ones estimated from COLMAP.
}\label{f5}
\end{figure*}

\begin{figure*}
  \centering
  \includegraphics[width = 6.5in]{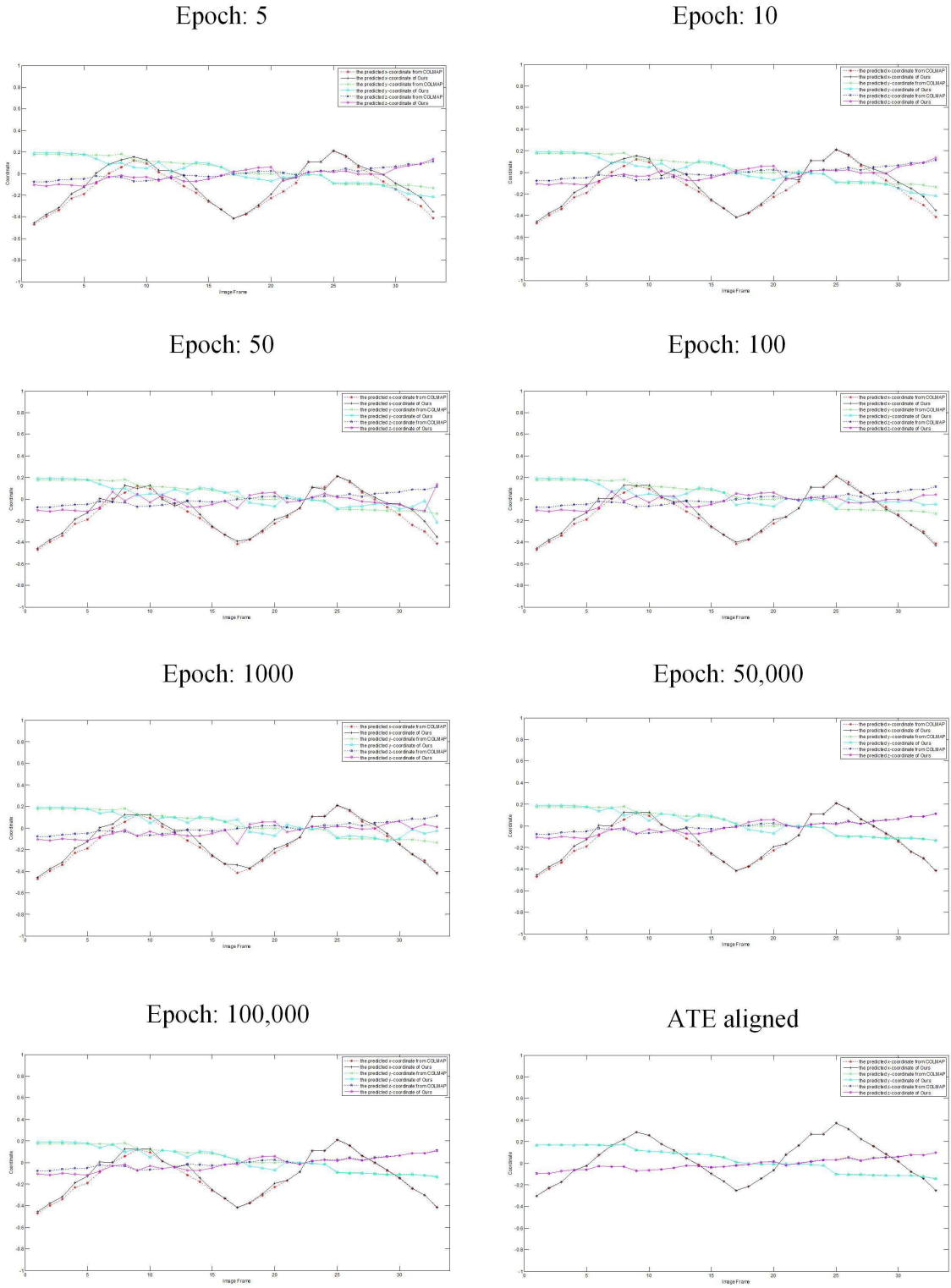}\\
  \caption{History of camera pose optimization during training. As each epoch goes by, our optimized camera translations gradually converge towards COLMAP estimations.
}\label{f6}
\end{figure*}

\begin{table*}[!htb]
\renewcommand\arraystretch{1.2}
\footnotesize
\begin{center}
\begin{tabular}{p{3.2cm}|p{1.0cm}p{1.0cm}p{1.0cm}p{1.0cm}p{1.0cm}p{1.0cm}p{1.0cm}p{1.0cm}}
\shline
Methods & \emph{Room} & \emph{Fern} & \emph{Leaves} & \emph{Fortress} & \emph{Orchids} & \emph{Flower} & \emph{T-Rex} & \emph{Horns} \\
\hline
BARF \cite{Lin2021} ICCV21	& 0.270	& 0.192	& 0.249	& 0.364	& 0.404	& 0.224	& 0.720	& 0.222 \\
NeRF-\,- \cite{Wang2021} arXiv21 &	\textbf{0.013}	& 0.007	& 0.006	& 0.041	& 0.018	& 0.011	& \textbf{0.013}	& 0.015 \\
Ours	& 0.019581	& \textbf{0.005418}	& \textbf{0.001678}	& \textbf{0.002514}	& \textbf{0.001096}	& \textbf{0.000788}	& 0.069863	& \textbf{0.001357} \\
\shline

\end{tabular}
\end{center}
\caption{Quantitative evaluation of our optimized camera translations on LLFF-NeRF dataset measured by ATE.}
\end{table*}

\section{Conclusion and Future Work}
Our approach allows for novel view synthesis without poses camera. The joint optimization improves the accuracy of both view synthesis and camera pose estimation by taking advantage of the merits of each other. The proposed approach eliminates the need for pre-computing the camera parameters using potentially erroneous SfM methods. Our experiments show that SaNeRF can effectively learn neural implicit scene representations and estimate the camera poses at the same time. However, our SaNeRF heavily relies on the SIFT matches; therefore, it is not suitable for novel view synthesis in the scenes with large texture-less regions, just like the synthetic dataset in \cite{Martinez2017}. On the other hand, as with other photometry-based reconstruction methods, it often struggles to reconstruct scenes with repeated structures. For example, the joint optimization struggles to converge on the \emph{scene 0316} from the ScanNet dataset. This is likely to be caused by the photometric ambiguity from the repeated structures. In the future, we plan on integrating the knowledge proposed in SfM/SLAM into SaNeRF to achieve robust novel view synthesis and self-supervised dense 3D reconstruction.

\section*{Acknowledgment}

This research is supported in part by the National Key R\&D Program of China (No. 2018AAA0102102).

\ifCLASSOPTIONcaptionsoff
  \newpage
\fi

\end{document}